\documentclass[journal]{IEEEtai}

\usepackage[colorlinks,urlcolor=blue,linkcolor=blue,citecolor=blue]{hyperref}

\usepackage{color,array}
\usepackage{array}
\usepackage{xcolor}
\usepackage{graphicx}
\usepackage{multirow}
\usepackage{cite}
\usepackage{amsmath,amssymb,amsfonts}
\usepackage{algorithmic}
\usepackage{graphicx}
\usepackage{textcomp}
\usepackage{tabularx}
\usepackage{booktabs}
\usepackage{tikz}
\usepackage{lipsum}

\usepackage{booktabs}
\usepackage{graphicx}
\usepackage{multirow}
\usepackage{subfiles}
\usepackage{xcolor}
\usepackage{xurl}
\usepackage{lscape}
\usepackage{hyperref}

\newcommand\copyrighttext{%
\footnotesize \textsuperscript{\textcopyright} 2024 IEEE.  Personal use of this material is permitted.  Permission from IEEE must be obtained for all other uses, in any current or future media, including reprinting/republishing this material for advertising or promotional purposes, creating new collective works, for resale or redistribution to servers or lists, or reuse of any copyrighted component of this work in other works. DOI: \href{https://www.doi.org/10.1109/TAI.2024.3351798}{10.1109/TAI.2024.3351798}}
\newcommand\copyrightnotice{%
\begin{tikzpicture}[remember picture,overlay]
\node[anchor=south,yshift=10pt] at (current page) {\fbox{\parbox{\dimexpr\textwidth-\fboxsep-\fboxrule\relax}{\copyrighttext}}};
\end{tikzpicture}%
}

\begin{document}
\copyrightnotice

\title{A Survey on Verification and Validation, Testing and Evaluations of Neurosymbolic Artificial Intelligence} 

\author{Justus Renkhoff$^{~*}$, Ke Feng$^{~*}$, Marc Meier-Doernberg, Alvaro Velasquez, and Houbing Herbert Song,~\IEEEmembership{Fellow,~IEEE}
\thanks{Manuscript received January 31, 2023. This work was supported in part by the U.S. National Science Foundation under Grant No. 2309760 and Grant No. 2317117.}
\thanks{Justus Renkhoff and Houbing Herbert Song are with the Security and Optimization for Networked Globe Laboratory (SONG Lab), Department of Information Systems, University of Maryland, Baltimore County, Baltimore, MD 21250 USA (e-mail: justusr1@umbc.edu; h.song@ieee.org).}
\thanks{Ke Feng is with the Department of Electrical Engineering and Computer Science, Embry-Riddle Aeronautical University, Daytona Beach, FL 32114 USA (e-mail: fengk2@my.erau.edu).}
\thanks{Marc Meier-Doernberg is with the Department of Electrical Engineering and Computer Science, Embry-Riddle Aeronautical University, Daytona Beach, FL 32114 USA (e-mail: meierdom@my.erau.edu).}
\thanks{Alvaro Velasquez is with the Department of Computer Science, University of Colorado, Boulder, CO 80309 USA (e-mail: alvaro.velasquez@colorado.edu).}
\thanks{*Justus Renkhoff and Ke Feng are co-first authors.}
%\thanks{This paragraph will include the Associate Editor who handled your paper.}
}

\markboth{}
{}

\maketitle

\begin{abstract}
Neurosymbolic artificial intelligence (AI) is an emerging branch of AI that combines the strengths of symbolic AI and sub-symbolic AI. Symbolic AI is based on the idea that intelligence can be represented using semantically meaningful symbolic rules and representations, while deep learning (DL), or sometimes called sub-symbolic AI, is based on the idea that intelligence emerges from the collective behavior of artificial neurons that are connected to each other. A major drawback of DL is that it acts as a ``black box", meaning that predictions are difficult to explain, making the {\color{black}testing \& evaluation (T\&E)} and validation \& verification (V\&V) processes of a system that uses sub-symbolic AI a challenge. Since neurosymbolic AI combines the advantages of both symbolic and sub-symbolic AI, this survey explores how neurosymbolic applications can ease the V\&V process. This survey considers two taxonomies of neurosymbolic AI, evaluates them, and analyzes which algorithms are commonly used as the symbolic and sub-symbolic components in current applications. Additionally, an overview of current techniques for the {\color{black}T\&E and V\&V processes of these components is} provided. Furthermore, it is investigated how the symbolic part is used for {\color{black}T\&E} and V\&V purposes in current neurosymbolic applications. Our research shows that neurosymbolic AI has great potential to ease the {\color{black}T\&E and V\&V processes} of sub-symbolic AI by leveraging the possibilities of symbolic AI. Additionally, the applicability of current {\color{black}T\&E and} V\&V methods to neurosymbolic AI is assessed, and how different neurosymbolic architectures can impact these methods is explored. It is found that current {\color{black}T\&E and} V\&V techniques are partly sufficient to {\color{black}test, evaluate,} verify, or validate the symbolic and sub-symbolic part of neurosymbolic applications independently, while some of them use approaches where current {\color{black}T\&E and} V\&V methods are not applicable by default, and adjustments or even new approaches are needed. Our research shows that there is great potential in using symbolic AI to {\color{black}test, evaluate,} verify, or validate the predictions of a sub-symbolic model, making neurosymbolic AI an interesting research direction for safe, secure, and trustworthy AI.

\end{abstract}

\begin{IEEEImpStatement}
Neurosymbolic AI allows the combination of symbolic representations or knowledge with the abstraction capabilities of sub-symbolic AI. This poses new challenges for the AI community, but also offers many new opportunities. As neurosymbolic AI is well suited for safety-critical domains such as autonomous systems, we aim to connect the two fields {\color{black}T\&E/}V\&V and neurosymbolic AI with our survey. Since neurosymbolic AI consists of several components, our research provides an overview of individual aspects regarding the {\color{black}T\&E/}V\&V of these components. Through this, we influence current research in the field of {\color{black}T\&E/}V\&V by highlighting opportunities as well as open challenges that emerge from neurosymbolic AI. Our research demonstrates that by combining symbolic and sub-symbolic AI, it is possible to {\color{black}test, evaluate,} verify and validate predictions made by {\color{black}non-transparent} sub-symbolic models. Accordingly, we provide an overview of current applications leveraging different architectures and combinations of symbolic and sub-symbolic AI, aiming to either test and evaluate in order to verify and validate predictions or to ease the T\&E/V\&V processes. In addition, the evaluation of current {\color{black}T\&E/}V\&V methods for their applicability to neurosymbolic applications revealed a need for testing frameworks that focus on neurosymbolic AI. As a result, we provide other researchers with possible directions for future research in the field of {\color{black}T\&E/}V\&V of neurosymbolic AI.
\end{IEEEImpStatement}


\section{Introduction}
% Step 1 - topic
\IEEEPARstart{N}{eurosymbolic} artificial intelligence (AI) is an increasingly important trend in machine learning (ML) and has been referred to as the 3rd wave of artificial intelligence \cite{garcez2020neurosymbolic}. The word “neuro” in its name implies the use of neural networks, especially deep learning (DL), which is sometimes also referred to as sub-symbolic AI. This technique is known for its powerful learning and abstraction ability, allowing models to find underlying patterns in large datasets or learn complex behaviors \cite{bishop1994neural}. On the other hand, “symbolic” refers to symbolic AI. It is based on the idea that intelligence can be represented using symbols like rules based on logic or other representations of knowledge \cite{Garnelo2019ReconcilingDL}. Neurosymbolic AI combines these two approaches to create a hybrid system that benefits from the reasoning abilities of symbolic AI and the adaptability of sub-symbolic AI, opening new opportunities to improve a variety of different AI branches \cite{10238788, velasquez2023darpa}.

% Step 2 - problem
A disadvantage of sub-symbolic AI is its nature of being a ``black box". This means that predictions made by these systems can be challenging to explain. Therefore, when an edge case leads to a system failure, it is often hard to find the reason for it. {\color{black} Accordingly, the rigorous testing \& evaluation (T\&E) and validation \& verification (V\&V) of these {\color{black}``black box"} is a relevant topic recognized by governments \cite{White_House_2023} and discussed in current literature \cite{10.1145/3459086.3459636, HUANG2020100270}.} As neurosymbolic systems incorporate a sub-symbolic component, this work aims to provide an overview of current techniques used to validate and verify the symbolic as well as sub-symbolic component, and how the architecture of neurosymbolic systems affects this process and can be used for V\&V purposes.

{\color{black} In software engineering, common terms are testing \& evaluation or T\&E and verification \& validation or V\&V. As defined by Wallace and Fujii in \cite{61456}, V\&V intends to ensure that software performs as intended and meets certain quality and reliability standards. T\&E are the methods and processes used to carry out V\&V. Validation refers to the process of ensuring that a system performs as expected and delivers the desired result with sufficient accuracy, while verification focuses on checking if the design and implementation is correct according to the specified requirements \cite{sargent1988tutorial}. Usually, verification is a process that takes place during development, while validation occurs at the end to evaluate if the program “does what it's supposed to do” \cite{28119}. For reasons of readability, we primarily use the term V\&V in the following.}

% Step 3/4 - Solution/Methodology

\begin{figure}
    \center
    \includegraphics[width=\linewidth]{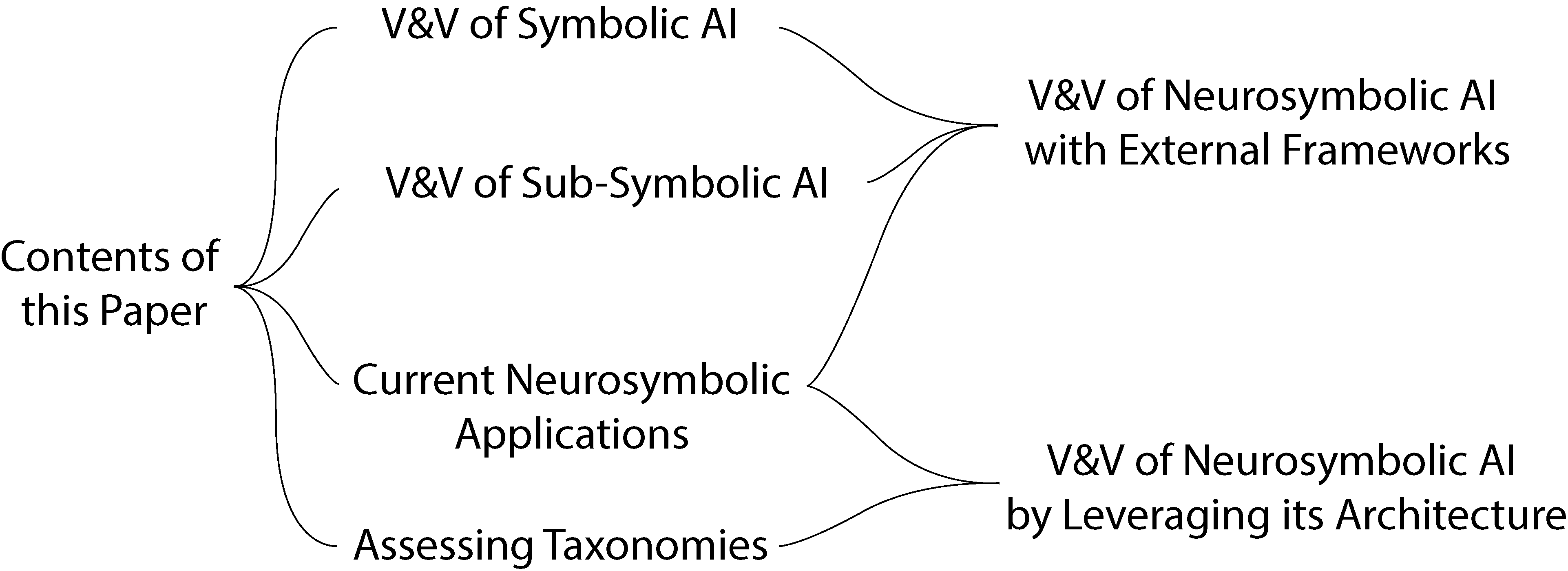}
    \caption{Contents of this paper.}
    \label{fig:contents}
\end{figure}

Recent frameworks propose methods to validate and verify symbolic and sub-symbolic AI, but discussing how the architecture of neurosymbolic AI can benefit the V\&V process of the system as a whole has not received enough attention yet. Therefore, this paper focuses on two areas. First, the concept of V\&V is mapped to symbolic and sub-symbolic AI, and an overview of current techniques and procedures used during the V\&V process is provided. Secondly, it assesses how different neurosymbolic applications use the symbolic side to enable V\&V of the sub-symbolic component. For this purpose, two different taxonomies of neurosymbolic AI are addressed, which categorize applications based on their architecture. 1) In 2020, Kautz proposed six possible designs of neurosymbolic systems \cite{Kautz_2022}. 2) An alternative taxonomy was introduced by {\color{black}Yu et al.} \cite{recent_advances} in 2021. These taxonomies are discussed and compared. Based on this, it is analyzed how current neurosymbolic applications leverage these architectures to use the symbolic component to make the sub-symbolic part more transparent, accurate, or safe, therefore enabling the V\&V process through a neurosymbolic system design. The structure of the discussion within this paper is visualized in Fig. \ref{fig:contents}.

% Step 5 - Result
{\color{black} Our work} demonstrates that some of the current testing methods used for V\&V are applicable to neurosymbolic AI. In particular, the combination of knowledge graphs (KGs) and DL is common, and it would be interesting to design a dedicated testing framework based on current techniques to validate neurosymbolic AI as a whole. However, there are also neurosymbolic AI applications that are not easy to test with current means. With this work, we show that there is much research potential in this area, and advocate the awareness of V\&V for neurosymbolic AI systems and AI in general.
Overall, this paper makes the following contributions:
\begin{itemize}
\item Present and compare two current taxonomies of neurosymbolic AI.
\item Map the concepts of V\&V as used in software engineering to symbolic and sub-symbolic AI.
\item Survey current V\&V approaches for symbolic and sub-symbolic AI.
\item Analyze the applicability of current V\&V methods to neurosymbolic applications.
\item Investigate how symbolic AI can support the V\&V process of sub-symbolic AI within a neurosymbolic system.
\item Discuss opportunities and challenges of V\&V in the domain of neurosymbolic AI.
\end{itemize}

\noindent The remainder of this paper is structured as follows: In section \ref{related_work} we analyze the related work. After that, in section \ref{categories_neurosymbolic} we examine and compare two different taxonomies for neurosymbolic AI. Then, in section \ref{validation} and \ref{sub-symbolic ai} we survey the most important methods to verify and validate symbolic AI and sub-symbolic AI respectively. In section \ref{opportunities} we analyze if these methods are applicable to current neurosymbolic AI applications and opportunities to leverage different neurosymbolic architectures using the symbolic part to verify and validate sub-symbolic AI. Afterward, in section \ref{open_challenges} we explain research gaps and problems that might be worth exploring in further research. In section \ref{conclusion} we summarize our findings and explain our planned future work.

\section{Related Work}
\label{related_work}

V\&V is a crucial process for ensuring the safety and reliability of safety-critical systems. Originally, V\&V processes were designed for conventional software without AI components. With the increasing number of modern applications utilizing AI, it becomes crucial to develop approaches for the V\&V of systems that use AI as a central element.

\subsection{Surveys on V\&V of Machine Learning}
V\&V of ML is an important topic in current research. For this reason, there are recent works and surveys that deal with this topic \cite{MLTesting, BRAIEK2020110542, 10.1145/3459086.3459636, HUANG2020100270, cardoso2021review}. Current surveys in this domain either deal with a specific area, such as autonomous systems \cite{cardoso2021review}, in which ML is used, or only investigate one aspect like ML testing \cite{MLTesting, BRAIEK2020110542} or formal verification of ML \cite{10.1145/3459086.3459636}, which are {\color{black}only parts} of the entire V\&V process.

In \cite{MLTesting}, testing of ML is surveyed. The survey presents current testing workflows, the components of an AI-based application that should be tested and provides an overview of properties that require testing as well as the frameworks that can be used to test these properties. Additionally, the survey showcases applications in safety-critical domains that need to be tested and how the testing workflows and frameworks can be applied to these applications.

Similar, in \cite{BRAIEK2020110542} testing approaches and current testing frameworks are presented. Compared to \cite{MLTesting}, this survey is not as extensive and does not provide background information about topics like ML in general which is covered in \cite{MLTesting}, but provides a comprehensive overview of current efforts regarding ML testing. 

Huang et al. \cite{HUANG2020100270} provide a detailed overview of verification, testing and the interpretability of DL within their survey. They define the terms verification and testing and explain the importance and meaning of properties like the robustness or interpretability of DL. They explain differences between current approaches for V\&V of DL and present a variety of testing frameworks and tools to increase the interpretability of DL.

\subsection{Surveys on Neurosymbolic AI}

There are multiple recent surveys that cover neurosymbolic AI and its applications in general \cite{garcez2022neural, sheth2023neurosymbolic, gaur2022knowledge, hitzler2022neuro, sarker2021neuro, susskind2021neuro, wang2022towards, gibaut2023neurosymbolic} and surveys that focus on more specific applications such as graph structures \cite{delong2023neurosymbolic}, biomedical knowledge graphs \cite{delong2023neurosymbolicbm}, or natural language processing \cite{hamilton2022neuro}. None of the just mentioned surveys covers testing, validation or verification in the domain of neurosymbolic AI and we could not find any surveys covering this topic to this date.

\section{Taxonomies of Neurosymbolic AI}
\label{categories_neurosymbolic}
Neurosymbolic AI covers a wide range of applications, and can be implemented in many different ways. This concerns on the one hand the selection of methods used on the symbolic side, and on the other hand how sub-symbolic methods are combined with the symbolic ones. Therefore, it is common to divide neurosymbolic AI into different categories. Accordingly, multiple taxonomies for neurosymbolic AI were proposed \cite{recent_advances, Kautz_2022, sheth2023neurosymbolic, susskind2021neuro}. In the following, two current taxonomies are discussed. The one from Kautz \cite{Kautz_2022} and Yu et al. \cite{recent_advances} are considered. Both taxonomies categorize neurosymbolic AI based on how the sub-symbolic and symbolic part interact with each other.
    
\subsection{Kautz's Taxonomy}
{\color{black} Currently, one of the most common categorizations is that of Kautz, who defines six different types of neurosymbolic AI  \cite{Kautz_2022}. All of these types represent different system architectures, that try to combine the advantages of symbolic AI with those of sub-symbolic AI.} Kautz defines the following categories:

\paragraph{Symbolic Neuro symbolic}
The input of the system is symbolic, then feed into a Neural Network, which outputs the symbolic result as well. A typical application of Symbolic Neuro Symbolic system is Natural Language Processing (NLP) and has become its Standard Operating Procedure (SOP) \cite{Kautz_2022}. The symbolic input are representing embeddings converted from a combination of words extracted from the original text document. There are a lot of approaches to perform this conversion, such as word2vec \cite{word2vec}, and Glove \cite{glove}. Then those symbolic inputs are {\color{black}fed} to a neural network that learns the underlying pattern to perform certain tasks, such as translation, semantic classification, and chat robot, etc. The output of the neural network is also symbolic in different forms based on the tasks. For example, the output is a sequence of words for translation tasks or a semantic label for classification tasks.

\paragraph{Symbolic[Neuro]} 
This type of neurosymbolic AI uses a symbolic approach as a problem solver in a neural pattern recognition subroutine. It is currently already being used in many fields. One of the best-known applications is AlphaGo Zero \cite{silver2017mastering}. Kautz states that most current autonomous vehicles and robots utilize this approach, but do not reference any applications from this domain.

\paragraph{Neuro$|$Symbolic}
The Neuro$|$Symbolic system performs symbolic reasoning based on non-symbolic input by leveraging neural networks to transform non-symbolic input (for example images) into a symbolic representation. The outputs of neural networks are fed into a symbolic representation which is used by a symbolic system to perform a complimentary task such as query answering \cite{garcez2020neurosymbolic }. All building blocks are connected so that learning happens in unison.
Garcez and Lamb \cite{garcez2020neurosymbolic} name the neuroßsymbolic concept learner \cite{mao2019nscl} and DeepProbLog \cite{manhaeve2018deepproblog} as examples.

\paragraph{Neuro: Symbolic $\xrightarrow{}$ Neuro}
Kautz describes this category as using the SOP which refers to the ``Symbolic Neuro symbolic" category. It has a special training regime based on symbolic rules. An example for this method is an application by Lample and Charton \cite{symbolic_math} which simplifies mathematical expressions. The description of this category is very abstract and Kautz refers to a formula for the training regime that is not explained further, which makes this category rather difficult to grasp.  

\paragraph{Neuro\_\{Symbolic\}}
Within this category, the symbolic part's purpose is to ``transform symbolic rules into templates for structures within the neural network" \cite{Kautz_2022}. Kautz references two examples{\color{black}, which are \cite{Serafini_neuro_symbolic} and \cite{symbolic_neuroSmolensky},} to show how this concept can be used to integrate abstraction and part-of hierarchies into neural networks.

\paragraph{Neuro[Symbolic]}
Neuro[symbolic] is inspired by the ``thinking fast and slow" theory from {\color{black}Kahneman \cite{kahneman2011thinking}, who} explains that the human brain has two different systems to make decisions. Neural Networks are similar to system 1, which operates automatically by instinct without control. The symbolic part is similar to system 2, which needs attention and effort to operate. Just like a human brain, most of the time system 1 is making decisions until it decides to invoke system 2 is necessary. A Neuro[symbolic] system relies on a neural network, and the embedded symbolic AI assists if invoked by the neural network. This type of neurosymbolic AI is considered to have the highest potential by Kautz\cite{Kautz_2022}. An example is a mouse-maze. Neural Networks recognize this task and invoke the symbolic engine, an algorithm to find the shortest path. The symbolic engine output the path with marks on the map which show the path. Then the neural network has been trained to interpret the marks and follow its guide to find the exit.
\\\\

Kautz's categorization demonstrates how the symbolic part cooperates with the sub-symbolic part of the application. This categorization is useful to understand how {\color{black}an application} as a whole works, but it also brings some problems with it. His categorization is very fine, and often it is difficult to clearly determine to which category an application belongs. Kautz explains some  categories only superficially and gives a few examples, which makes it difficult to understand certain categories thoroughly. While for other categories, there are no applications yet, so it is questionable whether they are at all useful in practice. In addition, the names of his category are not well-chosen. The categories, when pronounced, are sometimes impossible to tell apart and confusion can quickly arise. 
    
\subsection{Yu's Taxonomy}
{\color{black} Because of the critique on Kautz's survey, we present another survey by Yu et al. \cite{recent_advances} which provides an overview of current neurosymbolic applications and presents an alternative taxonomy.}
In their paper, current neurosymbolic AI applications are studied and divided into three groups: \textbf{learning for reasoning}, \textbf{reasoning for learning}, and \textbf{learning-reasoning}. Just like Kautz's taxonomy, the categories Yu et al. define represent how the symbolic part interacts with the sub-symbolic part of the application. In the following paragraphs, the taxonomy of Yu et al. will be explained shortly:

\subsubsection{Learning for Reasoning} 
\begin{figure}[h]
    \center
    \includegraphics[width=\linewidth]{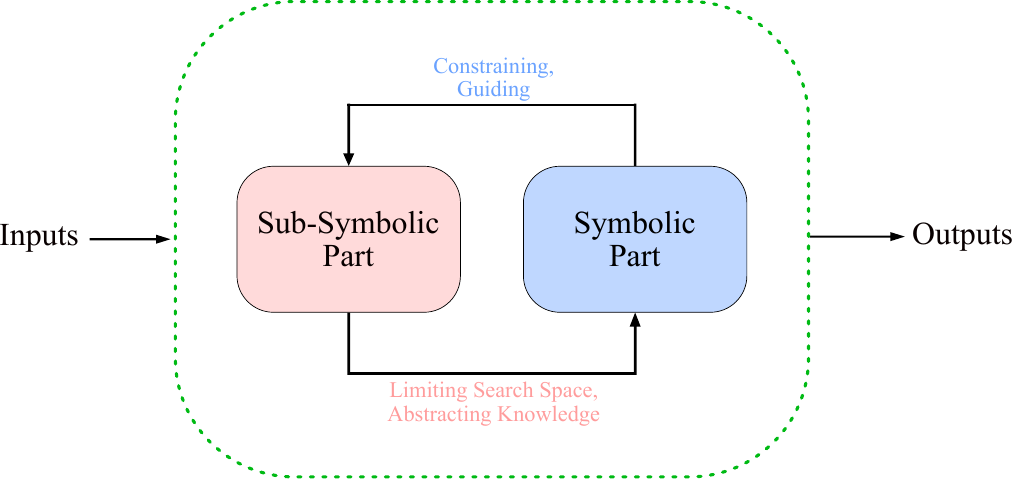}
    \caption{Flowchart of the Learning-Reasoning type of neurosymbolic AI. Here, the characteristics of the other architectures are combined and the two parts are in constant interaction \cite{recent_advances}.}
    \label{fig:Learning-Reasoning}
\end{figure}

{\color{black}This approach integrates sub-symbolic processes to enhance symbolic problem solving. Essentially, the sub-symbolic component narrows the search domain for the symbolic solver, optimizing the problem-solving process. This integration is depicted in Fig. \ref{fig:LearningForReasoning}.
\begin{figure}[h]
    \center
    \includegraphics[width=\linewidth]{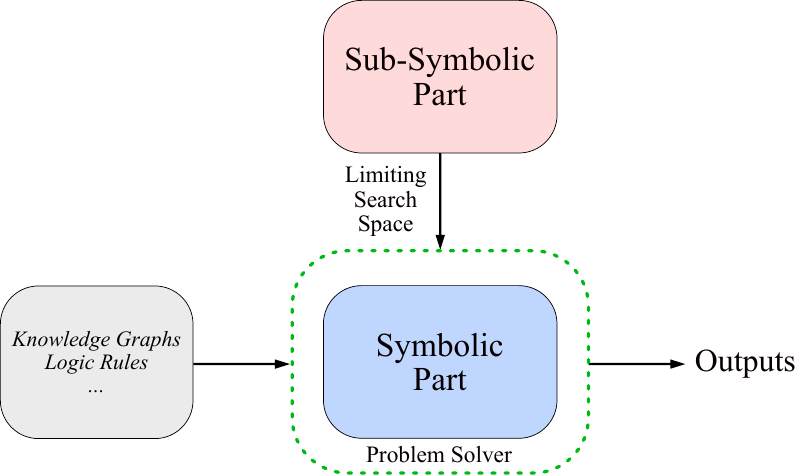}
    \caption{Flowchart of the Learning for Reasoning type of neurosymbolic AI. The sub-symbolic component is used to limit the search space for the symbolic part. Therefore, it is accelerating the process \cite{recent_advances}.}
    \label{fig:LearningForReasoning}
\end{figure}
Another way is that the sub-symbolic part converts unstructured data into symbols, to enable efficient symbolic reasoning as shown in Fig. \ref{fig:LearningForReasoning2}.
\begin{figure}[h]
    \center
    \includegraphics[width=\linewidth]{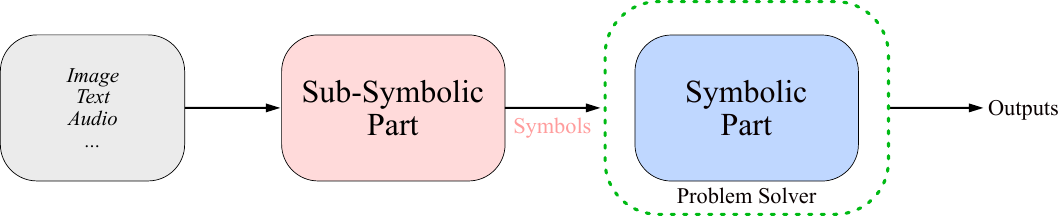}
    \caption{Flowchart of the Learning for Reasoning type of neurosymbolic AI. In this version of Learning for Reasoning, the sub-symbolic part transforms the knowledge that can be obtained from data to symbols \cite{recent_advances}. }
    \label{fig:LearningForReasoning2}
\end{figure}

\subsubsection{Reasoning for Learning} 
In this model, the roles are reversed: the sub-symbolic element primarily solves problems while the symbolic component supplements the neural network. This support manifests in two ways: firstly, by directing the neural network during its training phase, and secondly, by imposing constraints during prediction to prevent unsafe outcomes. Fig. \ref{fig:ReasoningForLearning} illustrates this architecture.
\begin{figure}[h]
    \center
    \includegraphics[width=\linewidth]{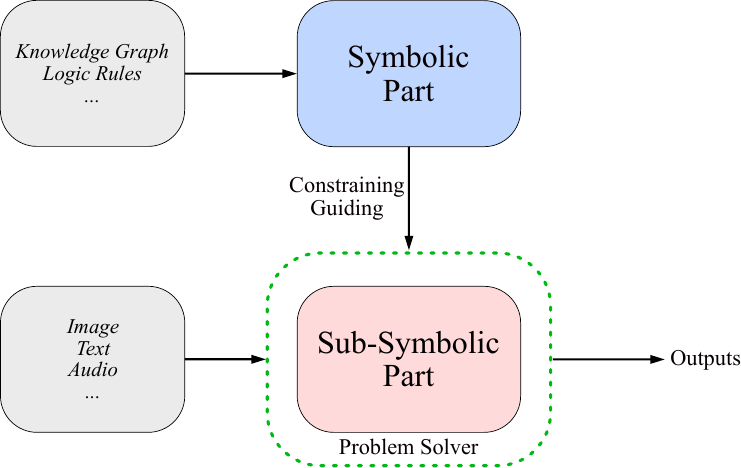}
    \caption{Flowchart of the Reasoning for Learning type of neurosymbolic AI. Here, the symbolic part can guide or constrain the sub-symbolic part \cite{recent_advances}.}
    \label{fig:ReasoningForLearning}
\end{figure}

\subsubsection{Learning-Reasoning} This variant represents a synergistic combination where symbolic and sub-symbolic elements collaborate equally in problem solving. Each component's output directly informs the other's input, creating a reciprocal and dynamic interaction. This bidirectional influence is visualized in Fig. \ref{fig:Learning-Reasoning}.}

\subsubsection{Example}
\begin{figure}[h]
    \center
    \includegraphics[width=\linewidth]{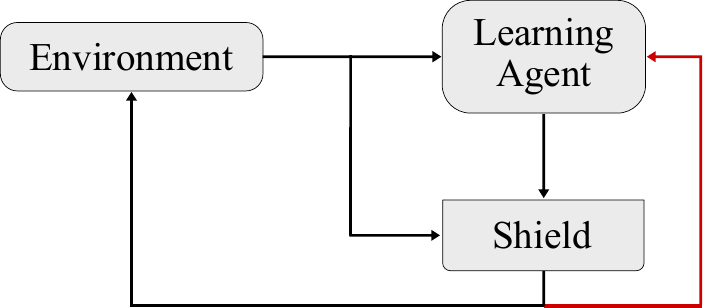}
    \caption{After the agents (sub-symbolic part) predicts an action based on the inputs from the environment, a safety shield (symbolic part) checks if this decision is safe and replaces it with a safe action if necessary. It is optional (indicated in red) to provide the agent with the information that the action was replaced or not \cite{alshiekh2017safe}.}
    \label{fig:post}
\end{figure}

Neurosymbolic approaches that implement safe reinforcement learning via shielding \cite{10.5555/3504035.3504361} are great examples to showcase this taxonomy as the shielding can be implemented in multiple ways. For a control task, a sub-symbolic model predicts an action, while a so called, safety shield, synthesized from safety specifications specified in temporal logic, ensures that every action is safe. If this application is implemented following the Learning for Reasoning or Learning-Reasoning design, first, the sub-symbolic part would make a decision based on its inputs from the environment. The decision is then given to the safety shield, that checks if the predicted action is safe and would then make minimal adjustments, if the action is determined to be unsafe. This concept would be categorized as Learning-Reasoning, in case feedback is provided to the sub-symbolic part, letting it know, that the action was replaced or not. If no feedback is provided, it would be Learning for Reasoning. This concept can be seen in Fig. \ref{fig:post}.

\begin{figure}[h]
    \center
    \includegraphics[width=\linewidth]{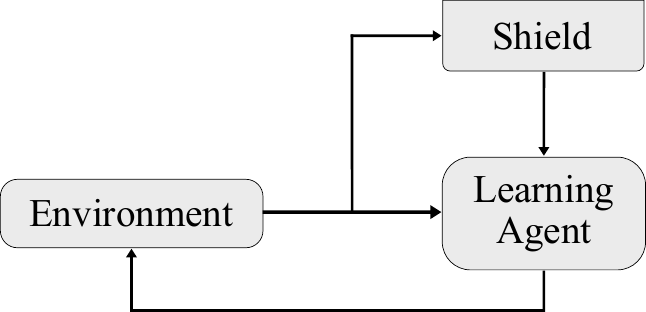}
    \caption{A safety shield (symbolic part) limits the actions the agent (sub-symbolic part) is able to choose from. Therefore, the agent is only able to choose from a set of safe actions \cite{alshiekh2017safe}.}
    \label{fig:pre}
\end{figure}

{\color{black} The work} \cite{alshiekh2017safe} is similar to \cite{10.5555/3504035.3504361}, but extends the paper by presenting an additional architecture in which the shield is inserted before the sub-symbolic part. This allows the shield to limit the action space to make sure that every action the sub-symbolic part can choose from is safe. This design would be ``Reasoning for Learning". This concept can be seen in Fig. \ref{fig:pre}.

In their survey, Yu et al.\cite{recent_advances} examine a wide range of current applications and classify them. They show that a variety of symbolic techniques can appear in every category of neurosymbolic AI. For example, first-order logic is used as a symbolic method in applications of all categories. This shows that the selection of the algorithms and methods for the symbolic as well as the sub-symbolic part is independent of the associated category. The categories in Yu's taxonomy are only based on the interaction of the symbolic and sub-symbolic component. In the following sections, we will analyze which frameworks and methods are currently used to test the most common symbolic and sub-symbolic methods and how the symbolic part of the application can contribute to the testing of the sub-symbolic component.

\section{V\&V of Symbolic AI}
\label{validation}
As a first step, the V\&V process of the two components of a neurosymbolic application are considered independently, with this section focusing on the symbolic component. Yu et al. \cite{recent_advances} shows that three methods in particular are used frequently as the symbolic part of a neurosymbolic AI system. These are propositional logic, first-order logic, and KGs. In the following section, the concepts of V\&V are mapped to these techniques and the {\color{black}capabilities to validate and verify} of symbolic AI are assessed.
\subsection{Mapping V\&V to Logical Systems}
In term of V\&V, the following properties are the most relevant: 1) Validity: A formula is valid if it is true under every possible interpretation or assignment. In logic, this means that if all premises of a statement are true, it is impossible that the conclusion is false. 2) Soundness: A logical system is sound if every statement that can be derived from the systems is true and an argument is sound if it is valid and its premises are true. 
\\
When mapping logical arguments to the V\&V process, the validity of a logic argument can be analogized to the verification phase. This is because a valid logical argument ensures structural correctness given that all premises are true, though it doesn't necessarily affirm the truth of those premises. This is similar to the verification process checking if the implementation or design of a system is correct according to its specification. On the other hand, soundness aligns with the validation phase, as an argument is deemed sound only when all its premises are unequivocally true. {\color{black}This is analog} to the validation phase, as this ensures that the output of a system is as expected and correct.

\subsection{Verification of Logic}

If a general algorithm can be found to prove the validity (true/false) of a logic argument, it is called decidable. Therefore, the {\color{black}question is}: Are propositional logic and first-order-logic decidable?

1) Propositional logic is decidable. The validity of a statement can be determined by a truth table. Truth tables are a fundamental process of computer science. As Anellis's research shows, it appears that this technique was used as early as the 19th century \cite{anellis2012peirce}. The complexity of this proof grows exponentially with the number of variables. Therefore, truth tables are in practice only usable for statements with a small number of propositional variables. Semantic tableau also called the truth tree method is an elegant alternative to truth tables \cite{Beth1955-BETSEA-10}. Accordingly, it is possible to validate the symbolic part of a neurosymbolic AI application that uses propositional logic using this technique, even if current standards are rather inefficient.

2) There is no general algorithm to check the validity of a first-order logic statement. Therefore, first-order logic statements are undecidable. However, this does not mean that it is impossible to show the validity of individual statements. Truth trees can be used to show the validity of first-order logic statements, but if the statement is invalid, the algorithm will run infinitely. To show an invalid statement, a countermodel has to be found.

As described above, algorithms have been found to check the validity of logic arguments. Even though first-order logic is not decidable, tools like \cite{githubGitHubWotpg}, can be used to verify it in many cases. However, validity does not depend on whether the premises are true. This means that the following statement would be valid in terms of logic:
\vspace{0.5em}
\textit{
\\ All animals are birds.
\\ All dogs are animals.
\\ Therefore, all dogs are birds.
}
\vspace{0.5em}\\
The above example shows that a statement can be valid but not sound. Soundness, as explained before, describes that not only the syntax but also the semantic is valid. Therefore, additional knowledge has to be used, to validate the semantic of such an argument. For this purpose, KGs could be used for validation purposes, which again have to be validated, too. This problem is considered in more detail in the following section.

\subsection{Validation of Knowledge Graphs}
KGs are an increasingly important component of current applications. Accordingly, there are numerous methods for validating these graphs. The survey ``Knowledge Graph Validation" by Huaman et al. gives an overview of current methods and tools \cite{knowledgeGraphVal}. Within this survey, several typical error sources which frequently occur in KG are described, and an overview is provided of which current tools are able to detect and correct such errors in order to create the most valid KG possible. %The error sources mentioned are \textbf{instance assertions}, \textbf{property value assertions} or \textbf{equality assertions}. Accordingly Huaman et al. list methods that tackle these problems. 
A broad variety of tools are available to validate KGs.
\subsubsection{Corroborative Fact Validation (COPAAL) \cite{syed2019copaal}}
To validate KGs or semantic statements, COPAAL computes a so-called mutual information (MI) score. The {\color{black}method} tries to find alternative sources on the web to validate a statement. The paper gives an example of how this method works; a given statement could be: ``Barack Obama is a US citizen". Using open databases and KGs such as DBpedia 2016-10\footnote{\url{https://www.faa.gov/air_traffic/publications/atpubs/atc_html/chap5_section_7.html}, accessed: 01/26/2023}, the method looks for similar statements that imply or refute that Barack Obama is a US citizen. E.g. the data could show that his place of birth is in the USA which would make it highly likely that he is a US citizen or if the method finds a source that states that he was a US President, it confirms that the original statement is very likely to be correct, giving it a high MI score. 

\subsubsection{Deep Fact Validation (DeFacto) \cite{lehmann2012defacto}}{
To validate knowledge, DeFacto is an algorithm that tries to find supporting information about a given fact in the information as well as supporting information from trustworthy sources. Additionally, it provides a score that represents the confidence DeFacto has when assessing the validity of a fact.

\subsubsection{Temporal Information Scoping (TISCO) \cite{rula2019tisco}}{
TISCO adds another component. This procedure tries to assign times to facts, since many assertions are only true at certain times. E.g. athletes regularly change their clubs, people may have different professions or live in different places at different points in their lives. Therefore, it is important not only to validate the facts, but also to link them to points in time in order to establish a timeline. 
}
\\\\
In addition to the above-mentioned procedures, there are other similar procedures with the same goal. In all procedures, different databases or the web are searched based on an assertion in order to confirm and validate statements. Other popular methods are FactCheck \cite{syed2018factcheck}, FacTify \cite{ercan2019retrieving}, Leopard \cite{speck2019leopard}, Surface \cite{padia2018surface} and S3K \cite{metzger2011s3k}.
Furthermore, there are already well build KGs available that are tested and highly validated, like YAGO\cite{rebele2016yago} or Conceptnet\cite{speer2017conceptnet}. YAGO is used by IBM in their Watson artificial intelligence system \cite{ferrucci2010building} and stores knowledge  about people, cities, countries, movies, and organizations. It was build with data from Wikipedia\footnote{\url{https://www.wikipedia.org/}, accessed: 01/29/2023}, WordNet\cite{10.1145/219717.219748}, which is also a widely used KG, and GeoNames\footnote{\url{https://www.geonames.org/}, accessed: 01/29/2023}. ConceptNet is a knowledge graph that links words and phrases with labeled edges. The information comes from a variety of sources, including crowdsourcing, expert-generated material, and games.
Another popular KG is DBpedia\footnote{\url{https://www.dbpedia.org/resources/knowledge-graphs/}, accessed: 01/26/2023} which builds on knowledge from Wikipedia documents.

\section{V\&V of Sub-Symbolic AI}
\label{sub-symbolic ai}
V\&V in the context of sub-symbolic AI is an exciting topic and also a big challenge, as deep learning is also often referred to as a ``black box" and is rather opaque in its decision-making. Accordingly, it is a challenge to validate and verify the behavior of these systems. In order to verify a system, it is usually checked whether certain requirements are met. This is usually done with the help of formal methods and is a current challenge for systems using sub-symbolic AI due to its complexity. For the validation of sub-symbolic AI different testing methods are used to check different properties like the correctness or robustness of a system.

\subsection{Verification of Sub-Symbolic AI}
In the survey of Huang et al. \cite{HUANG2020100270} various applications to verify sub-symbolic AI are presented. They provide a taxonomy for different verification approaches and define multiple properties that can be verified. In the following these approaches and properties as defined in \cite{HUANG2020100270} are summarized.
\subsubsection{Properties to Verify}
\paragraph{Robustness}\label{sec:robustness}Robustness can be defined as the ability of a model to make a correct decision even in situation when the input is noisy or manipulated \cite{drenkow2022systematic}.
\paragraph{Reachability \& Interval}{The reachability and interval are two very similar properties, closely connected to each other. {\color{black} Verifying} the reachability means, that  for a certain input the highest possible and lowest possible output is verified. {\color{black} Verifying} the interval is very similar, as it is an over-approximation of the reachability. 
}
\paragraph{Lipschitzian}{This property describes how the output changes when small changes are made to the input. When verifying this property, the change in output should remain below a specified distance.}

\subsubsection{Approaches}
\paragraph{Search-Based}{Verification algorithms belonging to this type verify the system through exhaustive searching. This approach uses algorithms such as the Monte-Carlo Tree Search for verification purposes \cite{wicker2018feature}.}
\paragraph{Constraint Solving}{Algorithms that leverage this approach convert neural network into constraints which are easier to verify because they are no longer a ``black box". For the verification of the resulting constraints, solvers like the SAT solver can be used.}
\paragraph{Over-Approximation} {Here, an over-approximation of possible outputs for an input is calculated for verification purposes. }
\paragraph{Global Optimization}{As the name suggests, these approaches are based on global optimization techniques. An example for this is the tool DeepGo \cite{ijcai2018p368} that uses global optimization techniques for verification in respect to reachability and robustness properties.}
\\\\ An in-depth {\color{black}explanation} for the verification of sub-symbolic AI can be found in \cite{HUANG2020100270}. {\color{black}Table \ref{tab:1} provides an overview of current approaches that can be used for the verification of sub-symbolic AI.} 

\begin{table}[h]
\centering
\caption{Approaches to Verify Sub-Symbolic AI as surveyed in \cite{HUANG2020100270}.}
\label{tab:1}
\begin{tabular}{l l} 
\toprule
\textbf{Approach} & \textbf{Publications} \\
\midrule
Search-Based & \cite{WU2020298,wicker2018feature} \\
\addlinespace
Constraint Solving & \cite{katz2017reluplex,ehlers2017formal,bunel2018unified,lomuscio2017approach,8318388,cheng2017maximum,narodytska2018verifying,ijcai2018p811} \\
\addlinespace
Over-Approximation  & \cite{gehr2018ai2,wang2018formal,raghunathan2018certified,ijcai2019p824,ijcai2018p368} \\
\addlinespace
Search-Based \& Constraint Solving & \cite{huang2017safety,dutta2017output} \\
\addlinespace
Over-Approximation  \& Constraint Solving & \cite{pulina2010abstraction,wong2018provable,mirman2018differentiable} \\
\addlinespace
Global Optimization & \cite{ijcai2019p824,ijcai2018p368} \\
\bottomrule
\end{tabular}
\end{table}

}

\subsection{Validation of Sub-Symbolic AI}
To validate sub-symbolic AI, a variety of measures can be tested. In \cite{MLTesting} these measures as well as the tools and frameworks to test these in order to validate such a system is surveyed. The measures addressed in this survey are the correctness, model relevance, efficiency, fairness, interpretability, privacy and robustness of the system. In \cite{HUANG2020100270} especially testing the robustness and increasing the interpretability are addressed. Depending on the use case of the application, some of these properties are particularly important. Within this survey especially the correctness, robustness and interpretability are considered for validation purposes. In the following, a brief overview of these measures and recent frameworks is provided.

\subsubsection{Properties to Validate}
\paragraph{Correctness}
Correctness is a fundamental property of a system, representing the probability that it completes a task correctly. Popular methods to measure the correctness are k-fold cross-validation \cite{kohavi1995study} and Bootstrapping \cite{efron1994introduction}. For classification tasks, metrics like accuracy, precision/recall, and ROC Curve are commonly used to measure the correctness. Suitability varies depending on the situation and data balance. Detailed examples can be found in Japkowicz's workshop \cite{MLEvaluation}. Regression problems can be evaluated using error measurements, such as Mean-Squared-Error (MSE) or Root Mean-Squared-Error (RMSE), which provide insights into expected deviations from the system's predictions. In summary, choosing the appropriate measurement is crucial to assess the correctness of a sub-symbolic system and should be carefully considered based on the task and data distribution. {\color{black} Table \ref{tab:2} shows a selection of works that focus on testing the correctness of sub-symbolic AI. It is based on applications surveyed in \cite{MLTesting}.}

\begin{table}
    \centering
    \caption{Works on Testing the Correctness as Surveyed in \cite{MLTesting}}
    \begin{tabular}{l l} % adjust the second column's width as needed
        \toprule
        \textbf{Testing Correctness} & \textbf{Publications} \\
        \midrule
        Testing Tools & \cite{10.1145/3183713.3196934, 10.1145/3077257.3077271} \\
        \addlinespace
        Testing the Input and Oracle Design & \cite{10.5555/3103620.3103631, nakajima2018generalized, dwarakanath2018identifying, 10.1145/3193977.3193980, breck2017ml, ma2018mode} \\
        \addlinespace
        Searching Data Bugs & \cite{hynes2017data, schelter2018automating} \\
        \bottomrule
    \end{tabular}
    \label{tab:2}
\end{table}

\paragraph{Robustness} The robustness property itself is similar to the one described in section \ref{sec:robustness}. The difference is that robustness can not only be verified, but also tested and therefore validated. The most common approach to test the robustness is to generate adversarial examples or inputs. Frameworks such as DeepXplore \cite{pei2017deepxplore}, DeepHunter \cite{xie2019deephunter} or DLFuzz \cite{guo2020coverage} use adversarial attacks to trigger misbehavior and therefore to test the robustness of a neural network. Techniques such as testing the code coverage, known from conventional software testing, can be adapted to sub-symbolic AI. These approaches maximize a metric called neuron coverage to improve the robustness of a sub-symbolic model. Another approach is to detect adversarial noise that might cause wrong predictions \cite{10.1145/3133956.3134057, tan2023noisecam}. While these methods focus on images, there are other approaches that focus on generating and detecting adversarial attacks for natural language processing \cite{QIU2022278} or cybersecurity \cite{9001114}, which can be used to test and improve the robustness of these models.

\subsubsection{Interpretability}
{Neural networks are often considered to be ``black boxes", because it is a challenge to comprehend the decision-making process of a trained model. However, in safety-critical and ethically sensitive domains, it is crucial to understand this process to prevent discrimination or system failures. Although there is no uniform definition of interpretability, previous work suggests that it refers to the degree to which humans can comprehend the reasoning and logic behind a deep learning system's decisions \cite{MLTesting, biran2017explanation}.

To evaluate interpretability, there are three main categories: Manual assessment, automatic assessment, and evaluation of interpretability improvement \cite{MLTesting}. Manual assessment involves humans in the loop and is evaluated in real applications. Automatic assessment, on the other hand, utilizes proxies to eliminate the need for human involvement. Identifying influential instances belongs to this approach, which can be achieved through two methods: Deletion Diagnostics and Influence Functions \cite{molnar2022}. Both methods detect influential instances by measuring the influence of the change to the model when modifying the data sets: Deletion Diagnostics remove data points, while Influence Functions up-weight instances by differentiating the loss function with respect to its parameters. Notable measures of Deletion Diagnostics are DFBETA\cite{ruiter2006national} and Cook’s distance\cite{cook1977detection}. 

\section{Opportunities}
\label{opportunities}
As shown in Fig. \ref{fig:oppotuities}, one solution is to verify and validate both sides of a neurosymbolic AI separately. Another solution is to leverage the characteristics of the symbolic AI to verify and validate sub-symbolic part. In the following, we will consider both approaches and assess whether and how current testing and validation methods can be applied to the isolated parts of a neurosymbolic application and how current applications leverage the characteristics of symbolic policies to validate or improve the properties of the sub-symbolic part.

\begin{figure}[h]
    \center
    \includegraphics[width=\linewidth]{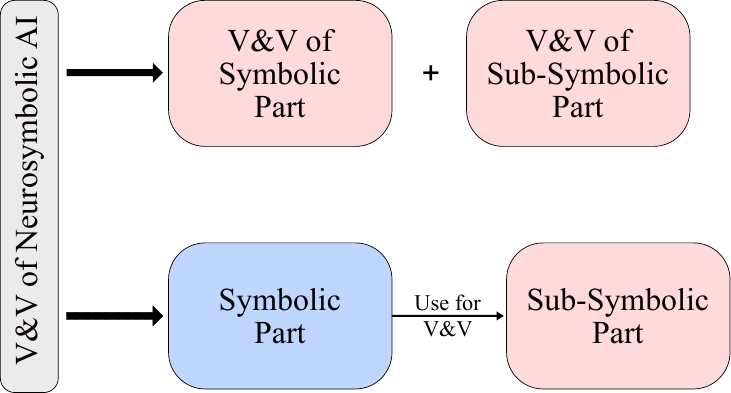}
    \caption{It is possible to either verify and validate the parts of a neurosymbolic application independently or the symbolic part can be used to either ease or conduct the V\&V process of the sub-symbolic part.}
    \label{fig:oppotuities}
\end{figure}

\subsection{Using Neurosymbolic System Architectures for V\&V}

Each of the three different categories of neurosymbolic AI defined by Yu et al. \cite{recent_advances} presented in their paper can affect the V\&V process differently. For example, in ``Reasoning for Learning", the symbolic part can support the sub-symbolic AI by providing guidelines and constraints through e.g. logic rules. This means that the input is directly applied to the sub-symbolic AI, as shown in Fig. \ref{fig:ReasoningForLearning}. The symbolic part can therefore improve the robustness and correctness of the system by checking, constraining or replacing decisions made by the sub-symbolic model. The category ``Learning for Reasoning" uses the symbolic part as problem solver. This means that the inputs directly go into the sub-symbolic part. It is feasible to transform the inputs to the sub-symbolic part to symbolic rules that allow to make transparent decisions and therefore increase the interpretability of the overall system. Both ``Learning for Reasoning" as well as ``Reasoning for Learning", have {\color{black}the} potential to improve the efficiency by accelerating the learning process either through guidance by symbolic rules or by limiting the search space with the sub-symbolic model. All of these concepts can also be applied to the category ``Learning-Reasoning". In the following, we give multiple examples based on current neurosymbolic applications that leverage these architectures and explain the opportunities these techniques provide to increase the safety and trustworthiness in AI and especially DL. An overview of selected applications is given in table \ref{tab:applications}.

\subsubsection{Safe Reinforcement Learning}
A popular application for neurosymbolic AI is safe reinforcement learning for autonomous control tasks. Alshiekh et al. \cite{10.5555/3504035.3504361} propose a concept to synthezise a safety shield from formal specifications represented in linear temporal logic. As already mentioned in section \ref{categories_neurosymbolic}, the integration of this shield into the neurosymbolic systems is very versatile and every architecture according to Yu's taxonomy is possible. In recent years, several similar approaches have been proposed, often using the ``Learning for Reasoning" or ``Learning-Reasoning" architecture to make the minimal needed adjustments to guarantee safe actions. One of the more recent works is ``Neurosymbolic Reinforcement Learning with Formally Verified Exploration" \cite{anderson2020neurosymbolic}. The paper introduces a reinforcement learning framework called REVEL. Similar to \cite{10.5555/3504035.3504361}, symbolic rules are used as a verification step within the deep reinforcement learning loop providing a safety shield that keeps the agent from executing unsafe actions. Therefore, this application can be categorized as ``Learning-Reasoning", showing how architecture can help to improve the correctness and robustness of a system. The paper demonstrates the results using a total of 10 benchmarks and compares them with similar state-of-the-art approaches. Compared to Deep Deterministic Policy Gradients (DDPG) \cite{lillicrap2015continuous}, the framework performs better in 7 out of 10 scenarios. Compared to Constrained policy optimization (CPO) \cite{achiam2017constrained}, however, it performs better in only 4 out of 10 cases. The survey ``A Review of Safe Reinforcement Learning: Methods, Theory and Applications" by Gu et al. \cite{gu2022review} provides an overview of safe reinforcement learning with many different approaches often using neurosymbolic AI for verification purposes. Additionally, the authors maintain a GitHub repository\footnote{\url{https://github.com/chauncygu/Safe-Reinforcement-Learning-Baselines}, accessed: 12/09/2023} listing current works in this domain.

\subsubsection{Verifiable Reinforcement Learning via Policy Extraction\cite{NEURIPS2018_e6d8545d}}
Another approach to increase the interpretability and transparency of the decisions of the sub-symbolic part of a neurosymbolic system is to derive rules from predictions. The neurosymbolic framework VIPER, which follows the ``Learning for Reasoning" architecture, is doing this by deriving rules from predictions of a neural network. These rules are represented by a decision tree. This approach helps to make decisions easier and more efficient to validate and verify. Furthermore, it makes the decisions of the entire system more transparent.

\begin{table}[t]
% \begin{adjustbox}{width=\columnwidth,center}
\caption{A selection of papers that use symbolic AI for V\&V of the sub-symbolic part}
\label{tab:applications}

\begin{tabular}{|p{0.5em}|p{1.5cm}|p{2.25cm}|p{2.25cm}|    }
% \begin{tabularx}{\columnwidth}{| X | X | X | X |}

\hline
\multicolumn{1}{|c|}{\textbf{Paper}} & \multicolumn{1}{c|}{\textbf{Category}} & \multicolumn{1}{c|}{\textbf{Summary}}                                                     & \multicolumn{1}{c|}{\textbf{V\&V Aspect}}                 \\ \hline
\cite{anderson2020neurosymbolic}     & Learning-Reasoning                     & Verify predictions using a symbolic ``safety policy"                                       & Increasing correctness and robustness by ensuring a safe output \\ \hline
\cite{tian2019learning}              & Non Applicable                                      & Convert 3D objects to code which is then converted to 3D shapes                           & Improves validity by increasing the interpretability        \\ \hline
\cite{NEURIPS2018_e6d8545d}          & Learning for Reasoning                 & Learn a provable decision-tree policy                                                     & Improves validity by increasing the interpretability        \\ \hline
\cite{trivedi2021learning}           & Learning for Reasoning                 & Learn programmatic policies from tasks that can be described by Markov Decision Processes & Improves validity by increasing the interpretability        \\ \hline
\cite{alshiekh2017safe}                    & Learning for Reasoning or Learning-Reasoning (based on implementation) & Restrict a DP model or overwrite its decision to allow safe reinforcement learning          & Validate predictions or restrict DL model to improve correctness and robustness           \\ \hline
\cite{huang2023laser}                    & Learning-Reasoning & Learning semantic video representations in a neurosymbolic weak supervised learning setup          & Verify learning results by checking against logical specifications         \\ \hline
\end{tabular}
% \end{adjustbox}
\end{table}
 
\subsubsection{Learning to Synthesize Programs as Interpretable and Generalizable Policies \cite{trivedi2021learning}}
This framework follows a similar approach as ``Verifiable Reinforcement Learning via Policy Extraction" \cite{NEURIPS2018_e6d8545d}. {\color{black}The difference} is that \cite{trivedi2021learning} does not use limited policy representations in the context of decision trees, but learns to synthesize a program soly on rewards. The derived policies can make the decisions more transparent than those of conventional DL methods.

\subsubsection{Learning to Infer and Execute 3D Shape Programs\cite{tian2019learning}}
This application is interesting because unlike the others, it does not quite fit Yu's taxonomy \cite{recent_advances} because it uses a total of two sub-symbolic parts and one symbolic part. Again, the symbolic part gives more accurate results and especially increases transparency compared to existing DL methods. 
The goal of the program is to represent a 3D object as 3D shapes. For this, first, an object is represented as code by means of a ``Neural Program Generator". Then a ``Neural Program Executor" converts the code to 3D shapes. The code is human-readable, and therefore it is possible to see which shapes of the 3D object have been recognized. This increases the interpretability. 

\subsubsection{LASER \cite{huang2023laser}}
{\color{black} Huang et al. \cite{huang2023laser} present a weakly supervised neurosymbolic learning approach to learn semantic video representations. The approach receives videos and spatio-temporal specifications in the form of linear temporal logic (LTL) as inputs. During the learning process an ``alignment score" of the specifications and the learned semantic representation is calculated. This allows for a verification of the learned representation. The alignment is optimized during the learning process. }

\subsection{Assessing the Applicability of Current T\&E/V\&V Methods to Neurosymbolic AI}
In this section, we address opportunities we have through current V\&V methods to determine where these approaches reach their limits in neurosymbolic applications. 

\subsubsection{Zero-shot Recognition via Semantic Embeddings and Knowledge Graphs \cite{Wang_2018_CVPR}} 
This method, which according to Yu's taxonomy \cite{recent_advances} belongs to the category ``{\color{black}Reasoning} for Learning", deals with zero-shot learning. The approach deals with unknown classes by using knowledge about previously learned classes and additional semantic embeddings. It uses both semantic embeddings and categorical relationships to predict the {\color{black}classes} of unknown pictures. The core of the application consists of two components: One component is a knowledge graph (KG), and the other is a graph convolutional network (GCN). The paper uses multiple configurations of datasets for its experiments. In the first one, the KG is based on relationships from Never-Ending Language Learning (NELL) \cite{carlson2010toward} and images are taken from the Never-Ending Image Learning (NEIL) \cite{chen2013neil} dataset. In the second configuration, the KG is based on the WordNet \cite{10.1145/219717.219748} database while the images for the GCN model's training are taken from the ImageNet \cite{russakovsky2015imagenet} dataset.
The KG can be validated with the previously analyzed methods. The paper investigates how the method behaves when noise is introduced in the KG and when it is completely random.  It is shown that the method is quite robust even when noise is present in the KG. However, if it is random, then the outputs are almost random guesses. Therefore, while it is important that the KG is validated by the GCN, which is the problem solver in this procedure, we compensate for noise but do not need to validate the KG perfectly and focus on the GCN. There are several types of the still rather new GCN. The type used in the paper is based on {\color{black}convolutional} neural networks (CNNs). This would mean that approaches such as \cite{zugner2019certifiable} to find robustness guarantees in GCNs could be used for verification and benchmarking tools like \cite{dwivedi2020benchmarkgnns, fung2021benchmarking} for validation purposes. Even though there are some works regarding V\&V of GCN it is a rather unexplored topic, which would be {\color{black}exciting} to further investigate.

\subsubsection{Alpha Go Zero \cite{silver2017mastering}}
AlphaGo Zero is an application developed by DeepMind. It is able to beat the world's best players in games like Chess or Go. This application is not listed in Yu's \cite{recent_advances} survey, but Kautz's \cite{Kautz_2022} uses it as an example for his category Symbolic[Neuro]. If this method were included in Yu's taxonomy, it would belong to the category ``Learning for Reasoning". A neural network evaluates the state of the game on the sub-symbolic part of the application, while a Monte Carlo Tree Search \cite{metropolis1949monte} tries to find the optimal move for the given situation on the symbolic part. Therefore, this application has a symbolic problem solver with a neural network supporting the decision-making process. AlphaGo Zero is trained by playing against itself in an attempt to find better moves and thus better models.
AlphaGo Zero's model trains itself and no human-generated data set is needed. This means that the system does not need to be protected against noisy or manipulated data. In addition, it is not a safety-relevant application. Therefore, robustness is not necessarily in the foreground, since targeted manipulations would be unlikely and futile. 
While the interpretability of AlphaGo Zero's decisions is interesting, it is not the priority when testing or verifying the system. The goal of AlphaGo Zero is to develop the strongest possible chess engine that can defeat any opponent. For this reason, the main focus in testing the program is on the correctness. To validate the Monte Carlo Tree Search, all possible moves for each possible game state would have to be evaluated to find the optimal solution. For games like TikTakToe, this would not be a problem, but since games like Chess or Go have too many different game states, this would not be feasible with current technology. To simplify this, the neural network looks at each game situation and evaluates it. Since numerous game states are very similar and similar moves would be optimal, the DL part tries to identify these relationships between the different situations to reduce the possibilities that need to be evaluated. This means that in order for the symbolic problem solver to be able to make the correct decision, the sub-symbolic part must have assessed the situation correctly beforehand. Thus, a labeled test data set would need to be created against which the model could be tested to make sure the game states are detected correctly. However, since AlphaGo Zero has never been beaten by a human, it is impossible to decide whether the human made a mistake in labeling or if AlphaGo Zero made a mistake in evaluating a game situation if differences occur. However, as improvements are constantly being made, especially in the efficiency of this process, it is clear that even though the models so far are very good, there is still room for improvement. The only way to test AlphaGo Zero at the moment is to let it play more games against itself to find better models, even if this is very inefficient. For this reason, however, optimizing the efficiency of AlphaGo Zero's training is also interesting, since the more efficient this is, the faster better results can be obtained and thus the most important property in this scenario, correctness, is also improved. To summarize this; there is no current test framework that would be applicable to an application like AlphaGo Zero.

\subsubsection{DeepProbLog\cite{deepproblog}}
DeepProbLog is a neurosymbolic AI framework belonging to the category learning for reasoning according to Yu's taxonomy. The sub-symbolic part is responsible for the low-level perception task, and the symbolic part then uses the learning result to perform logical inference. In their research, three sets of a total of six experiments are conducted to demonstrate the different abilities of DeepProbLog. In five out of six experiments, DeepProbLog outperforms the DL model itself, showing better generalization ability, less computational complexity and training time, and higher sample efficiency. The tasks in the experiments are the addition of single digits and multi-digits, sorting a list of numbers, and the coin-ball problem \cite{Coin-ball},  where the sub-symbolic part is used to recognize the numbers or colors in an image, and the symbolic part uses the classification results to complete the addition operation or to calculate the probability distribution.
The sub-symbolic {\color{black}components} used in these tasks are convolutional neural networks (CNNs) with basic architectures. The experiments used the MNIST data set. Input testing could be conducted to expose robusteness flaws \cite{MLTesting}. Testing {\color{black}frameworks} like DLFuzz could be used to generate adversarial samples and improve the robustness of the CNNs \cite{dlfuzz}. For the sorting task, the sub-symbolic part uses recurrent neural networks (RNNs) which are similar as the ones used in the work of Bosnjak et al. \cite{bovsnjak2017programming}. These could be tested by TensorFuzz \cite{tensorfuzz}, which is used to find undesired behaviors of RNNs. Also, cross-validation could be used during the training to validate the models performance. The symbolic part of DeepProbLog follows the inference process of ProbLog: First, generate the ground instances the query is based on; Second, rewrite the ground logic into a propositional logic formula; Next, the formula is compiled into a Sentential Decision Diagram (SDD) \cite{SDD} for more efficient evaluation; Finally, calculate the probability. Since this system is based on propositional logic, the symbolic part is decidable and could be verified e.g. using semantic tableau.

\section{Open Challenges}
\label{open_challenges}
Examining the current state of neurosymbolic AI and current V\&V methods, we have revealed numerous open challenges. These open challenges address neurosymbolic AI and its applications in general, as well as the  V\&V methods for both symbolic and sub-symbolic AI.
\paragraph{Investigating New Neurosymbolic Architectures}
The term ``neurosymbolic AI" is still relatively new at the time this paper was written. As our research has shown, it is difficult to find papers on the topic on the well-known platforms of ACM and IEEE. However, this is not because no one uses this concept, but because the term is not yet widely used in the scientific community. The works by Kautz \cite{Kautz_2022} and Yu et al. \cite{recent_advances} make important contributions by identifying and categorizing existing applications that use this technique. Similar works are published frequently, but there is still no widespread differentiation of different categories of neurosymbolic AI and terms as well as clear definitions must be established in the future. We have criticized Kautz's taxonomy for the fact that some of his categories are only theoretical with no applications implementing them and thus some categories are not practical at the moment. But this also shows that there are many opportunities to combine symbolic with sub-symbolic AI that have not been explored yet and are worth exploring to find out what potential neurosymbolic AI has.
\paragraph{Efficient Verification of Logic Rules}
Traditional methods like truth tables, which can be used to verify propositional logic, are very computationally intensive as their run-time depends on the number of parameters. This means that these methods do not scale well. However, depending on how the sub-symbolic part is related to the symbolic part, it may not be necessary to fully  verify the symbolic part. Neural networks have the advantage that they can usually deal well with noise in the data. That means, if the problem solver is the sub-symbolic part and the symbolic part has only a supporting function, it would be sufficient to approximate a complete verification. {\color{black} This approach could be further investigated and used to balance the computational cost and scalability with the need for accuracy and logical correctness.}
\paragraph{Testing of Emerging DL Architectures}
Methods for testing the correctness, robustness, and other metrics for neural networks are well-researched and are constantly being further developed. It happens again and again that new designs for neural networks are developed. These new architectures require either new testing methods or the adapting of existing ones. In the paper ``Zero-shot Recognition via Semantic Embedding and Knowledge Graphs" \cite{Wang_2018_CVPR} a GCN is used on the sub-symbolic part of the application. It would be interesting to explore whether it is possible to apply methods such as DLFuzz \cite{guo2020coverage} here.
\paragraph{Comparing the Efficiency of Neurosymbolic AI with Comparable Conventional Deep Learning Approaches}
Through neurosymbolic AI it is possible to perform the training process of a DL model in a more targeted way, since the symbolic part can guide and thereby support the sub-symbolic part during training and the decision-making process. Therefore, it would be interesting to compare whether neurosymbolic AI applications are more efficient in terms of runtime and possibly also in terms of energy consumption. {\color{black} Measuring the efficiency of software systems and AI are exciting topics that are currently being researched. Since energy-efficient training AI can save costs for companies and research institutions as well as protect the environment, it is exciting to look at the influence of neurosymbolic AI on the efficiency of training. The assessments could be based on existing metrics and test procedures for evaluating the resource efficiency of ML \cite{KERN2018199, guldner2022measuring}.} 
\paragraph{Apply Current V\&V Methods to Common Neurosymbolic Applications}
It could be tested whether existing V\&V methods can be applied to common neurosymbolic applications as explained in the opportunities area. The currently most popular neurosymbolic AI applications could be used as examples. This could be extended and a testing framework for neurosymbolic AI applications could be developed, because there are some configurations that are frequently used. For example, KGs are often combined with CNNs. Test frameworks could be developed for these standard configurations with respect to the architecture of the application.
{\color{black} \paragraph{Development of Dedicated Testing Frameworks for Applications using Neurosymbolic AI}
At present, there are only a few frameworks for testing neurosymbolic applications, as this is still a very new field.  While our paper focuses on testing the individual components and using symbolic AI to test the sub-symbolic component within the system, there are first frameworks that test the whole neurosymbolic system as such. These testing frameworks are showing initial success in domain-specific applications. For example, Large {\color{black}Language} Models (LLMs) are a popular area of application for neurosymbolic AI. Accordingly, the paper \cite{ngu2023diversity} introduces a ``diversity measure" based on entropy, Gini impurity, and centroid distance as a metric to determine the probability of failure of LLMs. Furthermore, for the neurosymbolic LASER \cite{huang2023laser} approach for learning semantic representations of videos a new model checker was needed. Accordingly, they implemented a model checker based on Scallop \cite{li2023scallop} for verification purposes. This shows that there is a great need for new model checkers and testing procedures for appplications based on neurosymbolic AI. }

\section{Conclusion}
\label{conclusion}
Within this paper, the current state of neurosymbolic AI was investigated, as well as the current possibilities to test, evaluate verify and validate neurosymbolic AI. Two taxonomies that categorize neurosymbolic applications based on the system's architecture describing how the symbolic and sub-symbolic parts of the application interact with each other were assessed. Afterwards, the standard procedures to verify and validate common approaches used on the symbolic as well as the sub-symbolic part were surveyed. Based on this, it was analyzed whether it is possible to apply these strategies to popular neurosymbolic applications mentioned in recent surveys. It was found that the applicability of current testing methods strongly relates to the algorithms used on the symbolic and sub-symbolic parts. While there are V\&V methods for most approaches used on the symbolic part, these are sometimes too computationally expensive for large-scale projects. Therefore, it is important to question how thorough the testing on this side has to be, since neural networks can handle noisy data well if the symbolic part is only supporting. For the sub-symbolic side, current testing frameworks can often be used for the V\&V. These may be modified if necessary, however, this area is still a vivid research area, and it may happen that neurosymbolic applications use concepts for which no current testing framework exist. Furthermore, some applications demonstrate how the symbolic part of the application can be used to make neural networks more transparent, robust or accurate. This approach offers many opportunities and is still very unexplored, so it is exciting to explore this technique in future works including different environments and applications. {\color{black} In addition, it was found that there is a growing need for dedicated testing frameworks specialized for domain-specific neurosymbolic applications. }

\bibliographystyle{IEEEtran}
\bibliography{ref.bib}

\begin{IEEEbiography}[{\includegraphics[width=1in,height=1.25in,clip,keepaspectratio]{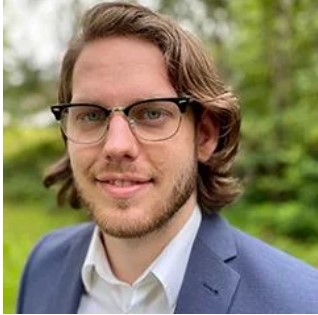}}]{Justus Renkhoff} earned a bachelor's and master's degrees in Media and Computer Science from Trier University of Applied Sciences, Germany in 2019 and 2021, respectively. He is currently pursuing a doctorate degree. Previously, he worked at the Institute for Software Systems at Trier University of Applied Sciences, the Security and Optimization for Networked Globe Laboratory (SONG Lab) at Embry-Riddle Aeronautical University and University of Maryland, Baltimore County and taught as an adjunct instructor at St. Bonaventure University, New York. His research focuses on explainable and neurosymbolic AI. 
\end{IEEEbiography}

\begin{IEEEbiography}
[{\includegraphics[width=1in,height=1.25in,,clip,keepaspectratio]{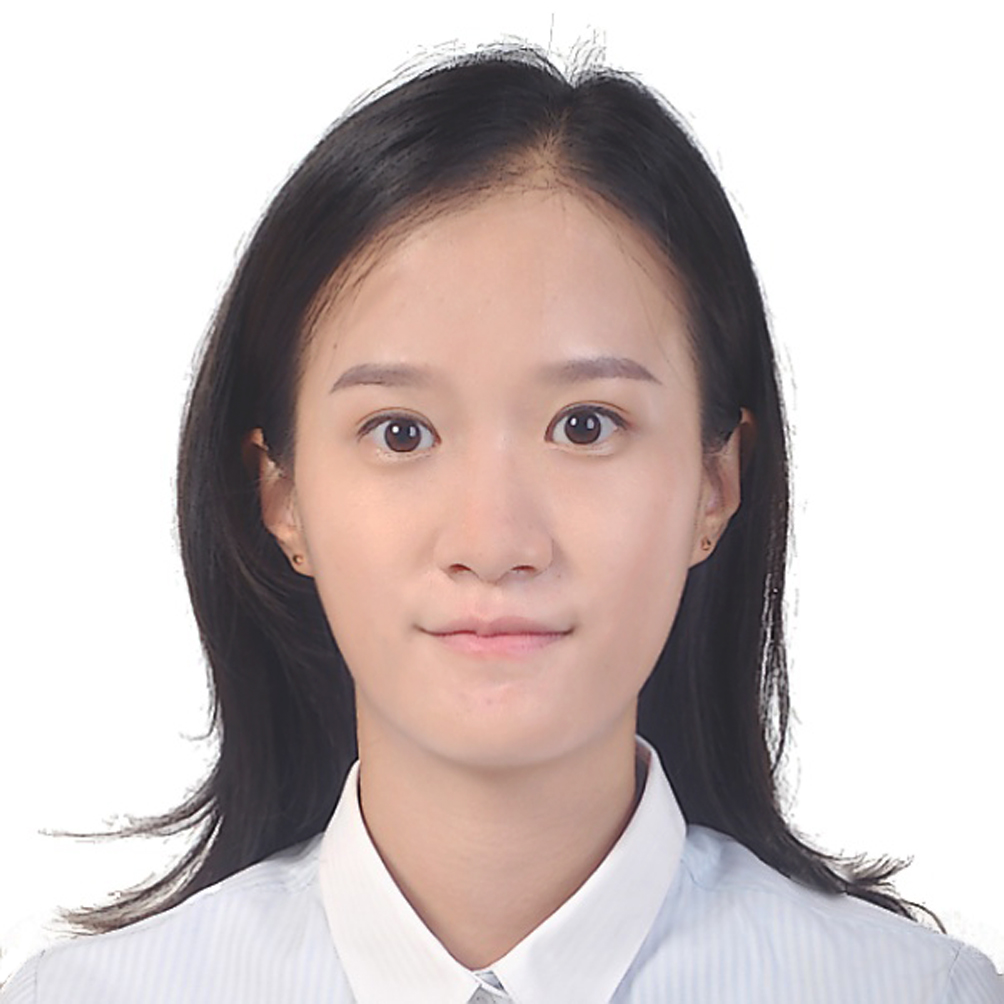}}]{Ke Feng} received master’s degree in Electrical and Computer Engineering from Embry-Riddle Aeronautical University(ERAU), Daytona Beach, Florida. She is currently pursuing a Ph.D. degree in Electrical Engineering and Computer Science at Security and Optimization for Networked Globe Laboratory, ERAU. Her major research interests include machine learning, deep learning, and the Internet of Things.
\end{IEEEbiography}

\begin{IEEEbiography}
[{\includegraphics[width=1in,height=1.25in,,clip,keepaspectratio]{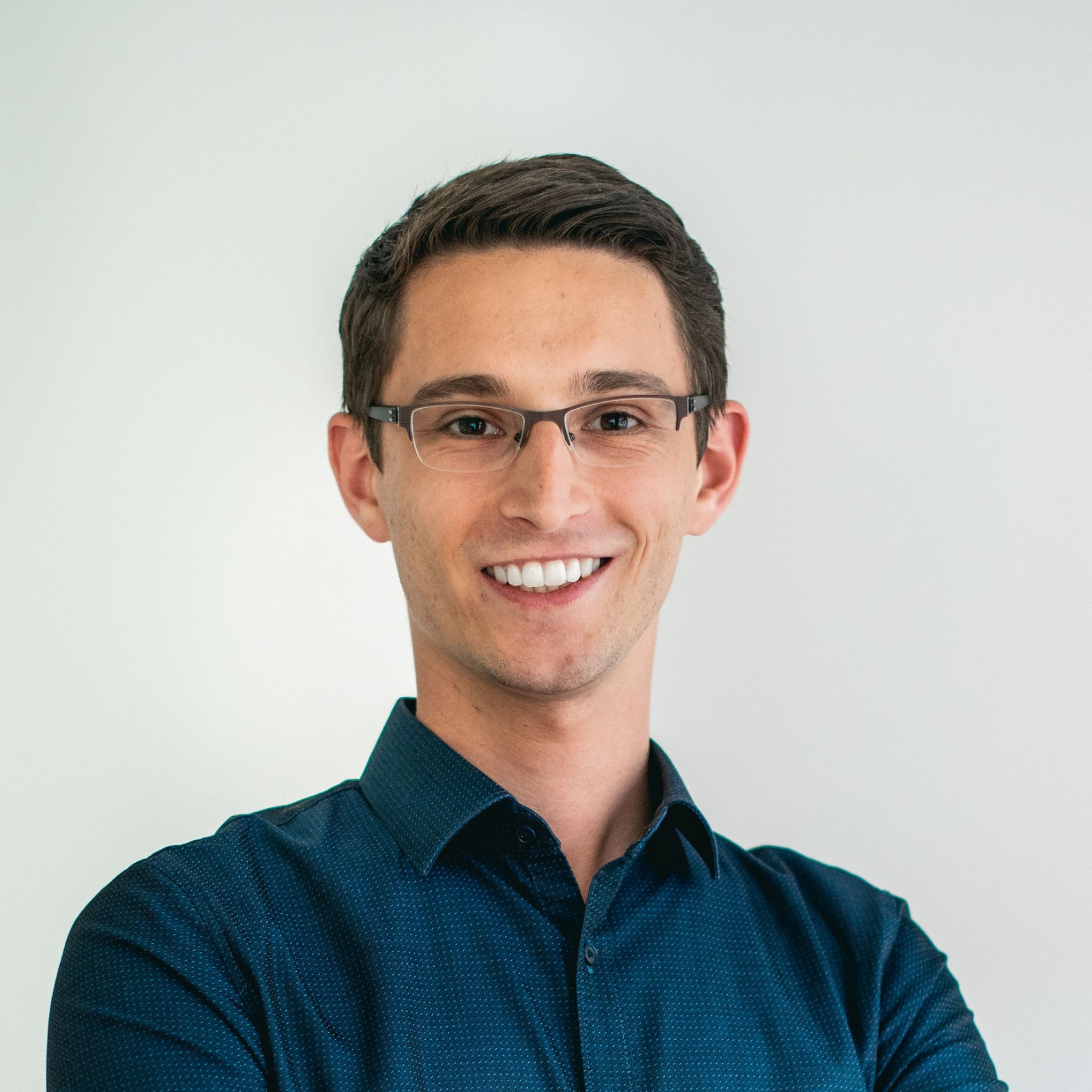}}]{Marc Meier-Doernberg} earned his master’s degree in Data Science from Embry-Riddle Aeronautical University, Daytona Beach, Florida. He currently works as a Lead Analyst for United Airlines where he develops data-driven approaches to air traffic management. He previously worked for Lufthansa Group and contributed to various analytics projects. He focuses on deep learning, machine learning, and their applications in aviation.
\end{IEEEbiography}

\begin{IEEEbiography}
[{\includegraphics[width=1in,height=1.25in,,clip,keepaspectratio]{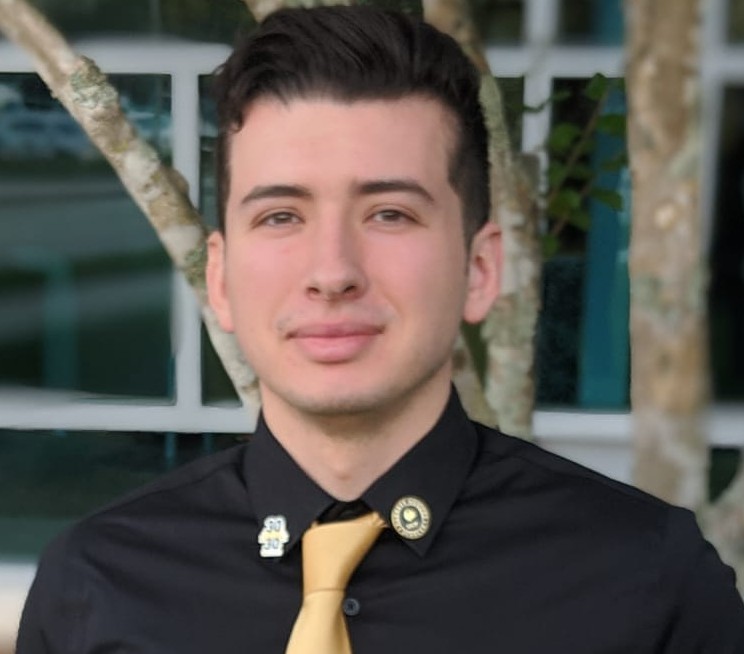}}]{Alvaro Velasquez} is a program manager in the Innovation Information Office (I2O) of the Defense Advanced Research Projects Agency (DARPA), where he currently leads the Assured Neuro-Symbolic Learning and Reasoning (ANSR) program. Before that, Alvaro oversaw the machine intelligence portfolio of investments for the Information Directorate of the Air Force Research Laboratory (AFRL). Alvaro received his PhD in Computer Science from the University of Central Florida and is a recipient of the National Science Foundation Graduate Research Fellowship Program (NSF GRFP) award, the University of Central Florida 30 Under 30 award, a distinguished paper award from AAAI, and best paper and patent awards from AFRL. He has co-authored 60 papers and two patents and serves as Associate Editor of the IEEE Transactions on Artificial Intelligence and his research has been funded by the Air Force Office of Scientific Research.
\end{IEEEbiography}

\begin{IEEEbiography}
[{\includegraphics[width=1in,height=1.25in,,clip,keepaspectratio]{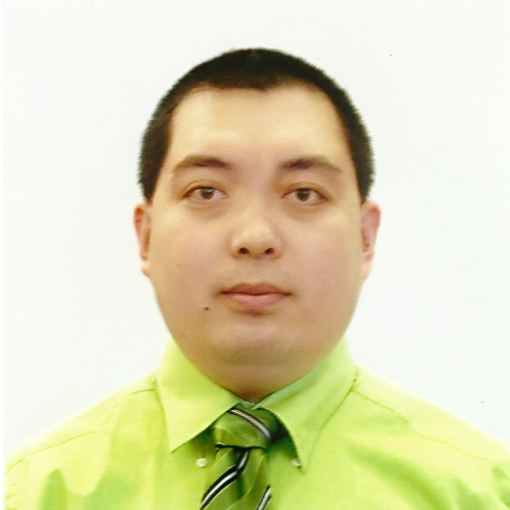}}]{Houbing Herbert Song} (M’12–SM’14-F’23) received the Ph.D. degree in electrical engineering from the University of Virginia, Charlottesville, VA, in August 2012.

He is currently a Professor, the Director of the NSF Center for Aviation Big Data Analytics (Planning), the Associate Director for Leadership of the DOT Transportation Cybersecurity Center for Advanced Research and Education (Tier 1 Center), and the Director of the Security and Optimization for Networked Globe Laboratory (SONG Lab, www.SONGLab.us), University of Maryland, Baltimore County (UMBC), Baltimore, MD. Prior to joining UMBC, he was a Tenured Associate Professor of Electrical Engineering and Computer Science at Embry-Riddle Aeronautical University, Daytona Beach, FL. He serves as an Associate Editor for IEEE Transactions on Artificial Intelligence (TAI) (2023-present), IEEE Internet of Things Journal (2020-present), IEEE Transactions on Intelligent Transportation Systems (2021-present), and IEEE Journal on Miniaturization for Air and Space Systems (J-MASS) (2020-present). He was an Associate Technical Editor for IEEE Communications Magazine (2017-2020). He is the editor of ten books, the author of more than 100 articles and the inventor of 2 patents. His research interests include cyber-physical systems/internet of things, cybersecurity and privacy, and AI/machine learning/big data analytics. His research has been sponsored by federal agencies (including National Science Foundation, National Aeronautics and Space Administration, US Department of Transportation, and Federal Aviation Administration, among others) and industry. His research has been featured by popular news media outlets, including IEEE GlobalSpec's Engineering360, Association for Uncrewed Vehicle Systems International (AUVSI), Security Magazine, CXOTech Magazine, Fox News, U.S. News \& World Report, The Washington Times, and New Atlas. 

Dr. Song is an IEEE Fellow (for contributions to big data analytics and integration of AI with Internet of Things), an Asia-Pacific Artificial Intelligence Association (AAIA) Fellow, an ACM Distinguished Member (for outstanding scientific contributions to computing), and a Full Member of Sigma Xi. Dr. Song has been a Highly Cited Researcher identified by Web of Science since 2021. He is an ACM Distinguished Speaker (2020-present), an IEEE Vehicular Technology Society (VTS) Distinguished Lecturer (2023-present) and an IEEE Systems Council Distinguished Lecturer (2023-present). Dr. Song received Research.com Rising Star of Science Award in 2022, 2021 Harry Rowe Mimno Award bestowed by IEEE Aerospace and Electronic Systems Society, and 10+ Best Paper Awards from major international conferences, including IEEE CPSCom-2019, IEEE ICII 2019, IEEE/AIAA ICNS 2019, IEEE CBDCom 2020, WASA 2020, AIAA/ IEEE DASC 2021, IEEE GLOBECOM 2021 and IEEE INFOCOM 2022.

\end{IEEEbiography}

\end{document}